\documentclass[sigconf]{acmart}

\AtBeginDocument{%
  \providecommand\BibTeX{{%
    \normalfont B\kern-0.5em{\scshape i\kern-0.25em b}\kern-0.8em\TeX}}}

\setcopyright{acmcopyright}
\copyrightyear{2018}
\acmYear{2018}
\acmDOI{10.1145/1122445.1122456}

\acmConference[Malta '20]{Malta '20: ACM Conference on the Foundations of Digital Games (FDG)}{September 15--18, 2018}{Malta}
\acmBooktitle{Malta '18: ACM Conference on the Foundations of Digital Games (FDG), September 15--18, 2020, Malta}
\acmPrice{15.00}
\acmISBN{978-1-4503-XXXX-X/18/06}

\newcommand{\Matthew}[1]{}
\begin{document}

\title{Tabletop Roleplaying Games as Procedural Content Generators}


\author{Matthew Guzdial}
\affiliation{%
  \institution{University of Alberta}
  \streetaddress{Address}
  \city{Edmonton}
  \country{Canada}}
\email{guzdial@ualberta.ca}

\author{Devi Acharya}
\affiliation{%
  \institution{University of California, Santa Cruz}
  \streetaddress{Address}
  \city{Santa Cruz}
  \country{United States}}
\email{dacharya@ucsc.edu}

\author{Max Kreminski}
\affiliation{%
  \institution{University of California, Santa Cruz}
  \streetaddress{Address}
  \city{Santa Cruz}
  \country{United States}}
\email{mkremins@ucsc.edu}

\author{Michael Cook}
\affiliation{%
  \institution{Queen Mary University of London}
  \streetaddress{Address}
  \city{London}
  \country{United Kingdom}}
\email{mike@possibilityspace.org}

\author{Mirjam Eladhari}
\affiliation{%
  \institution{S{\"o}dert{\"o}rn University}
  \streetaddress{Address}
  \city{Flemingsberg}
  \country{Sweden}}
\email{mirjam.palosaari.eladhari@sh.se}

\author{Antonios Liapis}
\affiliation{%
  \institution{University of Malta}
  \streetaddress{Address}
  \city{Msida}
  \country{Malta}}
\email{antonios.liapis@um.edu.mt}

\author{Anne Sullivan}
\affiliation{%
  \institution{Georgia Institute of Technology}
  \streetaddress{Address}
  \city{Atlanta}
  \country{United States}}
\email{unicorn@gatech.edu}

\renewcommand{\shortauthors}{Authors et al.}

\begin{abstract}
Tabletop roleplaying games (TTRPGs) and procedural content generators can both be understood as systems of rules for producing content.
In this paper, we argue that TTRPG design can usefully be viewed as procedural content generator design.
We present several case studies linking key concepts from PCG research -- including possibility spaces, expressive range analysis, and generative pipelines -- to key concepts in TTRPG design.
We then discuss the implications of these relationships and suggest directions for future work uniting research in TTRPGs and PCG.
\end{abstract}

%\begin{CCSXML}
%<ccs2012>
%<concept>
%<concept_id>10010405.10010469.10010474</concept_id>
%<concept_desc>Applied computing~Media arts</concept_desc>
%<concept_significance>500</concept_significance>
%</concept>
%</ccs2012>
%\end{CCSXML}

%\ccsdesc[500]{Applied computing~Media arts}



\maketitle

\section{Introduction}\label{sec:introduction}

Tabletop roleplaying games (TTRPGs) are a type of physical game similar to a board game, focused on players acting out particular roles. 
Arguably the most famous TTRPG is \emph{Dungeons \& Dragons} (\emph{D\&D}), an asymmetric game in which players take on roles as either adventurers or the ``dungeon master'' (DM) - a specific role for a player who poses challenges to the adventurers\cite{rpg_dnd}. 
However, there is a massive variety of TTRPGs, and a large variance in terms of the types of role (or roles) that players take on.

In recent years, there has been increasing interest in TTRPGs from game artificial intelligence researchers. 
This research ranges from work such as TTRPGs as a challenge for automated game playing agents \cite{martin2018dungeons}, to a rich space for AI-assisted tools \cite{liapis4,faria2019adaptive} and as a space for ontological exploration \cite{horswill2019imaginarium}. 
TTRPGs have also been a recent topic of AI-focused workshops \cite{liapis2018tabletopgamesworkshop,liapis2019tabletopgamesworkshop}.

In this paper we focus particularly on the overlap between TTRPGs and Procedural Content Generation (PCG).
We contend that the design of a TTRPG can be viewed as equivalent to the design of a procedural content generator.
A TTRPG can be understood as a generator that takes human players and authored mechanics as input and outputs play experiences. 
Frequently these play experiences take the form of narratives, for example a single \emph{D\&D} session telling the story of adventurers exploring a dungeon. 
In this scenario, the players' in-game choices directly build a story, driven by the interests of all players and the mechanics of the game they're playing.
The design of these mechanics confronts a TTRPG designer with many of the same questions as a PCG designer:

\begin{itemize}
\item How does the system incorporate random noise (structured random values) and what kind of noise does it draw on to ensure that the output varies without feeling arbitrary?
%MPE May12: perhaps add a tiny explanation/example of what noise can be, just in a parenthesis in the above bullet.
%MJG May 12: Sure!
\item How does the system ensure that the mechanics (process/functions of a PCG system) do not lead to a broken or frustrating experience?
\item How do players seed a generator with pieces of content (e.g. in \emph{D\&D} pieces such as non-player characters, puzzles, and monsters) and how can the system be designed to output a space of desired experience (e.g. in \emph{D\&D},  a play session \textit{feels} like an adventure)?
\end{itemize}

Because of these similarities, we contend that positioning TTRPG design as PCG design provides value to PCG researchers for the following reasons:

\begin{itemize}
\item  TTRPG design is a type of practical and common example of PCG design. This means we can study TTRPG designers as PCG designers, which allows for an additional vector of study into the design of content generation. 
We expect that this could lead to a broader understanding of PCG.
\item Because human players work with analog generation to produce stories in TTRPGs we can consider them parts of analog mixed-initiative generators \cite{liapis2016mixedinitiative}. 
This is valuable as such mixed-initiative generators with human and AI partners are still under-researched. 
Thus we expect that researching TTRPGs from this angle could improve our understanding of mixed-initiative systems broadly.
\item TTRPGs can be viewed as generative systems that are composed of several distinct components (e.g. fighting mechanics, character creation). 
We argue that there is value in looking at these systems through the lens of more traditional PCG approaches. 
We expect that these components create rich areas to apply PCG as there is a history of analogical generation of adventure scenarios dating back to the 1970's \cite{smith2015analog}.  %AL: cite computer-generated dungeon here, from the 80s, see Gillian's Analog History of PCG paper.
%\item As players are part of the generative system in TTRPGs, this leads to unique and complex challenges for automated play. %AS: I'm with Antonios below. I did an attempt with the first line, but I'm not sure how Juliani's work fits into this
%The PCG researchers will be well-suited to aid such efforts as in \cite{juliani2019obstacle} only if they have a strong understanding of the domain. %AL: not sure what this "means" (or, well, what it wants to argue), actually. Perhaps you should unpack it.
%MJG May 12: Yup, looking at this again I'm not sold on this. I broke apart the first two points and kept the last one
\end{itemize}

In this paper, we delve into the question of why TTRPG design is a valuable domain for PCG researchers. 
We start by formally defining TTRPGs through the lens of PCG and present additional connections between general PCG concepts and TTRPGs. 
We then discuss how current facets of PCG research relate to TTRPGs, using examples to clarify our points. 
Finally, we discuss how viewing TTRPGs as PCG can open additional avenues for future work.

\section{A TTRPG Primer}\label{sec:ttrpg_intro}%Assigned: Matthew and Antonios
%AS - this section is billed as being for the PCG audience, so would be good to tie back to PCG in a few places, even though the next section makes it more explicit
TTRPGs are social activities, exercises in imagination, usually played in a shared physical or digital space by a group of human \textbf{players}.
In many TTRPGs, each player controls a \textbf{Player Character} (PC) that usually collaborates with other PCs towards a joint goal, such as completing a quest, solving a mystery, surviving, etc.
Each TTRPG comes with sets of rules, guidelines, and mechanics that provide a framework for  creative play. 
Many TTRPGs also include mechanics that use dice or cards to introduce constrained randomness that impacts the outcomes of the PCs' actions. 
TTRPG play typically looks like players around a table or digital space, engaging in ``in character'' (while playing the game, as the PC) and ``out of character'' (as themselves) conversation, centered on the game world, followed by in-game actions and the particular game's resolution mechanic (to see whether those actions succeed or fail). %\todo{SOMETHING MISSING HERE?}
%AS - maybe some examples of how play works?
%creates adventures for the characters, who navigate its hazards and decide which paths to explore.
%MJG May 12: I like the idea, I'm not sure how to give examples without bringing up a specific system. Maybe give examples of kinds of play from a few systems? Or say something about a "prototypical" play experience?
%MJG May 16: Added the final sentence here

In many TTRPGs, one of the participants acts as the lead storyteller and referee, who creates adventures for the characters, determines the results of the adventurers’ actions, and narrates what they experience. 
This participant's role falls somewhere between player and designer, and as such is sometimes not even identified as a player. Instead, this participant is named ``Dungeon Master'', ``Storyteller'', ``Game Master'' among others; this paper will use the term \textbf{Game Master} (GM) for this role. 
In games like \emph{Dungeons \& Dragons}, a GM usually controls all non-player characters (NPCs), carrying out their interactions (e.g. dialog) with the PCs or with other NPCs. 
This creates an asymmetrical experience, where players decide on the paths laid out by the GM.
However, across all TTRPGs there is a great deal of variance in terms of which players control what elements of the game world. 
For example, in \emph{Legacy: Life Among the Ashes} \cite{rpg_legacy} each player controls both a specific PC and the ``family'' of that PC, making choices that will influence how that family changes across generations.

While the most well-known TTRPGs have a GM who controls the themes, story arcs, and moment-to-moment dice resolutions, there exist a class of games referred to as ``GMless'' or ``GMfull'' in which all players share equal agency in impacting the storyworld. 
These games may have players controlling one character only, acting out events as in improv theater, e.g. in \emph{Fiasco} \cite{rpg_fiasco}, or may have players act more like a GM or direct storyteller, determining the world and actions of many characters within it, e.g. in \emph{The Quiet Year} \cite{rpg_quietyear}. 
Usually games do not fall cleanly into either of these two extremes of play.
For instance, players not involved in a scene in \emph{Fiasco} impact its high-level outcome (success or failure), thus acting in part as a collective (and rotating) GM.
The distribution among players of creative authority over different parts of the storyworld is an active area of innovation in modern TTRPG design~\cite{BuildingWorldsTogether}.

Given that the majority of tabletop RPGs have no digital component, it is typical to include random number generation in order to keep outcomes surprising for players. 
Randomness in this instance ``adds drama, it breaks symmetry, it provides simulation value, and it can be used to foster strategy through statistical analysis'' \cite{costikyan2013uncertainty}. 
Especially in TTRPGs, tension builds up as players pick up the physical dice, and when they commit to an action that has a high degree of risk \cite{costikyan2013uncertainty}.
Without some degree of randomness, one could imagine a player who could just say ``I hit it and it dies'' to end every fight. 

Frequently randomness is facilitated by rolling dice, where the goal is to roll above some threshold to succeed, but other random number generators are also possible. 
For example, \emph{Spindlewheel} \cite{rpg_spindlewheel} uses a custom deck of Tarot-like cards to determine if a player's action succeeds and to what extent; these sources of noise are referred to as {Oracles}, as their output must be ``interpreted'' by the participants in the game according to the game's mechanics.
TTRPGs with progression systems, which are also common in digital games, will have characters level up in ways that bias these probability distributions in the player's favor. 

We can consider the types of players and how they resolve actions in the storyworld as components of a generator. 
The players act as agents, working in and against the confines of the games world to produce the experience of play. 
As the players are the primary drivers of this experience, we next discuss relationships between players and their characters and between different players.

While the roles of players and GM and the way they play the game are fairly clear-cut, an important aspect of TTRPGs is the relationships that emerge between players in the group and between players and their characters. 
% relationship between player and character
At the core of the RPG experience is that of playing a role, and different players take on this task in different ways. 
Bowman categorizes the relationships between characters and the players who created them into nine types, based on the ``sameness''  between a player's primary identity and their character \cite{bowman2010functions}. 
Examples include the Fragmented Self which augments or twists one normally minor part of a player's personality, or the Oppositional Self which embodies behaviors the player may find abhorrent in daily life \cite{bowman2010functions}. 
The character concept may begin through external inspiration, conversations with the group, or the player themselves, but the player's relationship with their PC evolves over the course of the stories told around the table, and based on the other players' responses.

% relationship between players / or players and GM
Beyond their own character, players also influence each others' actions and emotions tremendously during the game session, and group dynamics are important to keep track of.
Players who choose to be disruptive (e.g. causing harm to other players' characters) can swiftly ruin the co-operative effort around the table of telling a good story and are sometimes dealt with harshly by a GM \cite{wick2015playdirty}.
Beyond bad behavior, a roleplaying group is likely to include different types of players who prefer different problems and have different ways of solving them. 
Robin D. Laws identified seven player types -- including the Power Gamer, who optimizes their character to take the most out of the game's rules, and the Method Actor, who prefers resolving ethical dilemmas based on their character's carefully fleshed out psychological profile \cite{robin2002laws}.

In a game run by a GM, identifying players' different priorities and providing opportunities for each of them to shine is a major aspect of session management. 
On the grander scale, players have preferences in terms of genre, e.g. modern, horror, science-fiction, fantasy, and further categorizations within those broad genres. Identifying the genre and themes that the group wants to explore, the intended duration of the game, as well as limits, sensitivities and taboo themes, requires that the group is in constant and honest communication both before and during a TTRPG. 
As players' relationships with their characters evolve during the game, so do the group's dynamics, their preferred themes or decision-making styles; these changes affect the type of stories told during the play experience.

%AS - this is repeating what is above, maybe cut or move? Is the information about one-shots/campaigns and modules helpful for the PCG audience? 
%MJH May 12: I think knowing that some players play pre-written modules is helpful for a PCG audience as it's a pretty natural "oh we could generate those" sort of problem. What if we say that explicitly?
Tabletop roleplaying games are frequently either played once in \textbf{one-shot} sessions or in longer forms called \textbf{campaigns}. 
In campaigns, the same group of players meets regularly, with the GM guiding the players through some longer adventure and the other players playing the same characters and progressing these characters as they play. 
There are existing pre-written adventures called \textbf{modules}, especially popular among players of \emph{D\&D}, but it is also common for players to invent their own worlds, characters, and adventures and explore them week-to-week. 
We expect that modules for a TTRPG would be an excellent generation challenge for PCG researchers, who would have to balance generating a branching narrative and appropriate combat and puzzles for that narrative.

\section{TTRPGs under a PCG Lens}\label{sec:ttrpg_pcg_lens}

In this paper we view TTRPGs as hugely complex procedural content generators, systems composed of many moving parts and sub-generators. These components include the players of the game, who bring their individual interests, personalities, and storytelling predispositions.
Also integral to this system is everything previously established in the world of the game (either in previous sessions or pre-authored in a module) and the particular mechanics of the games.
Together these components create a near unlimited but biased space of possible output stories. 
Each component matters in this system.
For example, absent the players, the mechanics of \emph{D\&D} suggest a high fantasy story like the \emph{Lord of the Rings} \cite{tolkienLOTR} or \emph{Game of Thrones} \cite{martinGOT}. 
However, players impact this system as well, and so there are those who play \emph{D\&D} in science fiction, mystery, and superhero settings, adapting the mechanics to fit their interests.
We lack the space and expertise to more formally break down the ways in which player psychology impacts the generation process; thus we instead focus on the other elements of this system for the remainder of this paper.
%AL: unnecessary controversy (the part in parentehsis below, which I removed), also note there are already books published on the psychology of RPGs (we also doubt such expertise to fully understand a player exists), 
%MJG May 12: Haha fair!

Content generators are often grouped according to their outputs, such as the Game Bits, Game Space, and Game Scenarios framework proposed by Hendrikx et al. \cite{hendrikx2013procedural}. 
In this framing we can describe TTRPGs as generators that typically output a story. 
In this paper, when we talk about story, narrative, and  storytelling (narration), we  conform, for the sake of simplicity, to Genette's theoretical framework for narrative analysis \cite{genette_narrative_1983}. 
By \textbf{story}, we mean a  temporal sequence of events.  
A \textbf{narrative} is a story the way it is told. Narration concerns {how} to tell a story, e.g. the art of \textbf{storytelling}. 
We recognize that the design space in TTRPGs is one of narrative potential \cite{laurel_placeholder:_1994}. 
However, we note that TTRPGs do not always or do not exclusively produce stories. 
For example, \emph{The Quiet Year} outputs both the history of a small town over the course of a year and a map of that town \cite{rpg_quietyear}.
TTRPGs do not have to produce stories at all; the game \emph{Oh No Bro, You Thought Too Hard About The Bangers And You Fell Into The Banger Singularity} produces a playlist but explicitly no narrative \cite{rpg_bangers}.
At the far end of this spectrum, there is a class of TTRPGs referred to as Lyric Games which draw on poetry and essay forms to produce games which are only meant to be read \cite{lyricGames}. 
Thus, the process of reading these games becomes itself a form of play, an exercise of imagination of what playing them would be like, and their only output is this experience. 
For scope and complexity reasons, we primarily focus on TTRPGs that output stories in this paper, in other words TTRPGs that we can consider story generators \cite{kybartas2016survey}.

We note that the framing of ``games output stories'' may apply to games broadly.
For example we can draw comparisons between how chess games were retold as poetic stories such as ``To The Lady That Scorned Her Lover'' by Henry Howard \cite{yalom2004birth}, recent TTRPGs that produce stories via a game of chess like Takuma Okada's \emph{Chess: Two Kingdoms} \cite{rpg_chess}, and modern chess-like video games like \emph{Fire Emblem} (Nintendo, 1990) that lead to player stories and inspire fanfiction \cite{zagal2009ethically}.
However, for reasons of scope we focus on TTRPGs in this paper.

\subsection{TTRPG Mechanics as PCG Approaches}

Different approaches to PCG produce different outputs. 
Despite the complexity of a TTRPG when viewed as a story generator, given the inherent complexity of an arbitrary group of human players, different decisions made during the design process can bias the kinds of stories the game tends to output. 
In particular, we focus on the designed mechanics of a TTRPG and how they can be viewed as representations of distinct PCG approaches. 

The concept of ``Flavors of Noise'' appears in PCG design: what kind of random noise one might use as the basis for a generator (e.g. Perlin, Simplex, etc.) \cite{shaker2016noise}.
The same concept appears in TTRPG design, most frequently in games with dice mechanics \cite{costikyan2013uncertainty}. 
For example, most actions in \emph{D\&D} are resolved (i.e. determined to have failed/succeeded) based on rolling a 20-sided die. 
This means that any player has an equal chance of rolling a critical fail (rolling a 1) and a critical success (rolling a 20). %; this offers a somewhat slapstick situations, with a PC succeeding at a high-skill action only to immediately fail a simple one. %AL: this is not (exactly) true, that's what Difficulty Classes are for. There is no equal chance of failing a simple task as a complex one. But explaining this would take a lot of space and is not really important for the message.
%MJG May 12: So it's a fairly frequently reported thing to have this kind of big turnaround in D&D (from failure->success or success->failure). See: (https://www.reddit.com/r/DnD/comments/5tgs1y/5e_from_dangerous_to_slapstick_how_my_party_went/) I agree with you that it's slightly more complicated than I made it out, with modifiers and various other things trying to combat it, but I don't think the details really matter?
Comparatively, the Powered by the Apocalypse (PbtA) system resolves most actions by rolling two 6-sided dice, with a resolution system where rolling 6 or less indicates failure, a roll of 7 to 9 indicating a partial success (or success at some cost), and a roll of 10 or more indicating a complete success. 
Because the probability distribution of two dice is centered on 7, the most common result is for a PC to only barely succeed, which is appropriate for the original PbtA game, \emph{Apocalypse World}, in which players played survivors just barely getting by in a post-apocalyptic wasteland \cite{rpg_apocalypseWorld}.

One of the most common methods for story generation within PCG research and practice is through some sort of grammar \cite{kybartas2016survey}. 
Eschewing dice and other kinds of explicit randomness, the \emph{Belonging Outside Belonging} (BOB) TTRPG system uses what can be thought of as a grammar \cite{rpg_belonging}. 
In this game system, PCs have weak, regular, and strong moves. 
Weak moves give the player a token when used and strong moves cost a token to use. 
These moves play off one another, acting as grammar rules, which can be applied in sequence to produce stories.
A weak move can be used at any time but produces a problem in the narrative and gives the player a token. 
A strong move has a precondition that the player must have a token and produces a solution to a problem. 
For example in the game \emph{Dream Askew} \cite{rpg_belonging} the PC class ``The Iris'' has a weak move ``Draw unwanted attention to your movements'' and a strong move ``Get out of harm's way''\footnote{\url{https://buriedwithoutceremony.com/wp-content/uploads/2019/11/Dream-Askew-Play-Kit.pdf}}.
Thus a player playing The Iris can use the first move and then get out of trouble with the second move.
Just as a grammar for natural language generation strings together words or sentences based on rules, the BOB system strings together player actions to produce narrative.
%AL: not sure why this is a grammar; may need unpacking.
%MJG May 12: Made an attempt at this, any better?

\section{Relevant TTRPG Facets for PCG Research}\label{sec:facets}

We have made the argument for TTRPGs as procedural content generators, describing them as complex systems of players, game histories, and particular game mechanics. 
In this section we identify a number of open areas of PCG research and how they relate to TTRPGs, in order to make the argument about the value of TTRPGs to PCG research. 
In some cases, the facet will represent a novel vector for PCG research and in others, it will represent how one might study existing PCG research topics in TTRPGs.
% Not super familiar with PCG terminology, but for someone with more knowledge of this, the kinds of limits on generation here can be more compared to how PCG systems/generators work 

% https://dtwelves.com/gaming/safety-calibration-cards/
% Lines and Veils forum discussion: http://indie-rpgs.com/archive/index.php?&topic=12904.msg138054#msg138054
% Lines and Veils stack exchange discussion: https://rpg.stackexchange.com/questions/30906/what-do-the-terms-lines-and-veils-mean
% ^ The above cites Sex and Sorcery as the origin of the use of Lines and Veils. Does anyone have a copy of this supplement? 

\subsection{Possibility Space and Safety Tools}\label{sec:facets_possibilityspace}%Assigned to: Devi

The concept of possibility or generative space represents the theoretical space of all possible output of a generator. 
Being able to impact this space, sometimes referred to as the ``controllability'' problem, is a core research problem for PCG~\cite{smelik2010integrating,yannakakis2011experiencedriven}. %AL: citations needed here, especially on controllability
%MK: added a couple of controllability cites
Recent approaches such as Danesh \cite{cook2016danesh} seek to empower users of PCG generators to more directly alter this space.

Tabletop roleplaying games use specific practices, designs, and mechanics to shape the possibility space of play. 
This shaping of the possibility space can help to provide boundaries for the world, limiting what can occur during play. 
We have already touched on how TTRPG designers shape these boundaries ahead of time through the choice in mechanics.
Boundaries can also be shaped during play by integrating setting-specific limitations into the game mechanics.
For instance, as part of the history creation RPG \emph{Microscope} \cite{robbins2011microscope}, during setup players are asked to define a ``palette'' of things they do and do not want to see during play. 
The palette consists of two columns, ``Yes'' and ``No'', into which players add ``ingredients'' that will either belong or not belong in the created history. 
Each player then may add one ingredient to the palette in rounds, with players continuing to add to the palette until a round where a player chooses to not add anything else. 
The ``Yes'' column is reserved for elements that players would like to see in the world that might be unexpected given the setting of the game, but which players agree can show up during play. 
Conversely, the ``No'' column is used for elements that might be expected to be brought up in the history given the setting, but that players want banned; these elements will not be brought up in the history generation of the game.
This may be because a player is uninterested in a particular topic or due to a negative or traumatic association with the topic. 
This style of directly allowing players to draw boundaries and taboo themes across the possibility space of a particular TTRPG session are referred to broadly as \textbf{safety tools}.

Other safety tools are agnostic to the game system, i.e. general rules that are agreed upon by players before starting play. 
For instance, before starting a new game, GMs might hold a ``session 0'' where players can establish pregame agreements for the course of the game. %AL: cite session 0
%MJG May 12: Did you have something in mind for a citation here?
One such example is establishing lines and veils, developed by Ron Edwards \cite{rpg_sorcerer} as a way for players to individually determine the limits for what they want to see in the game. 
Lines represent defined limits for the game that the player does not want to see crossed--things that will not occur at all in the events of the game. 
Veils represent ways of still allowing for certain themes without putting them front-and-center--they represent a fade-to-black moment or actions that might still occur in the background. 
For example, a scene with two characters who are romantically involved fading to black before any explicit content. 
By defining these before the actual game starts, players can help to shape what will and will not occur in the game, and what forms these themes and actions take. 

An example of changing the space of play during the game itself is the X-Card, a safety tool created by John Stavropoulos in which the GM places a card with a large X in the center of the table \cite{stavropoulosxcard}. 
If players are uncomfortable or would like to shift the way that the story is going during play, they can tap or raise the X-Card and all players drop the current storytelling thread. 

We note that these approaches are not fool-proof and that safety tools are only part of shaping a culture of safety at a table. 
This is important as in a TTRPGs the interplay of the players and mechanics can lead to broaching topics that no one individually could have anticipated coming up. 
We can think of this as being caused due to the complexity of a TTRPG understood as a story generator.

%AL: missing here how all these options for defining the boundaries of possibility space can be implemented (or have been implemented in PCG)

%MK: took a pass at this, feel free to revise. maybe this belongs at the start of this section, alongside the other PCG cites?
Similarly, declarative constraint-based approaches to PCG -- such as answer set programming (ASP) -- attempt to explicitly provide users with tools for shaping the possibility space to exclude undesirable outcomes.
ASP's ``integrity constraints'' can be used to describe potential properties that the user does not want generated artifacts to have, and prevent artifacts with these properties from being generated~\cite{ASPforPCG}.
Integrity constraints have been widely applied to this problem in PCG -- for instance in the ASP-based abstract game generator Gemini~\cite{Gemini}, which permits the user to provide the generator with a ``design intent'' containing integrity constraints that block off undesirable parts of the full Gemini possibility space.
However, we note that generally the developers/designers of a PCG generator are the only ones with access to shaping the generator output like this, and that it is unusual to allow for such direct shaping of a generator's output after development/during use.
One example of this is Horswill's Imaginarium tool~\cite{horswill2019imaginarium} which applies a declarative constraint-based PCG approach to the generation of entities (such as monsters, NPCs, and treasure items) for TTRPG encounters, with the user able to specify constraints that explicitly forbid generator outputs with undesirable properties.

\subsection{Expressive Range}\label{sec:facets_expressiverange}%Assigned to: Matthew

\begin{figure}[tb]
  \centering
  \includegraphics[width=3in]{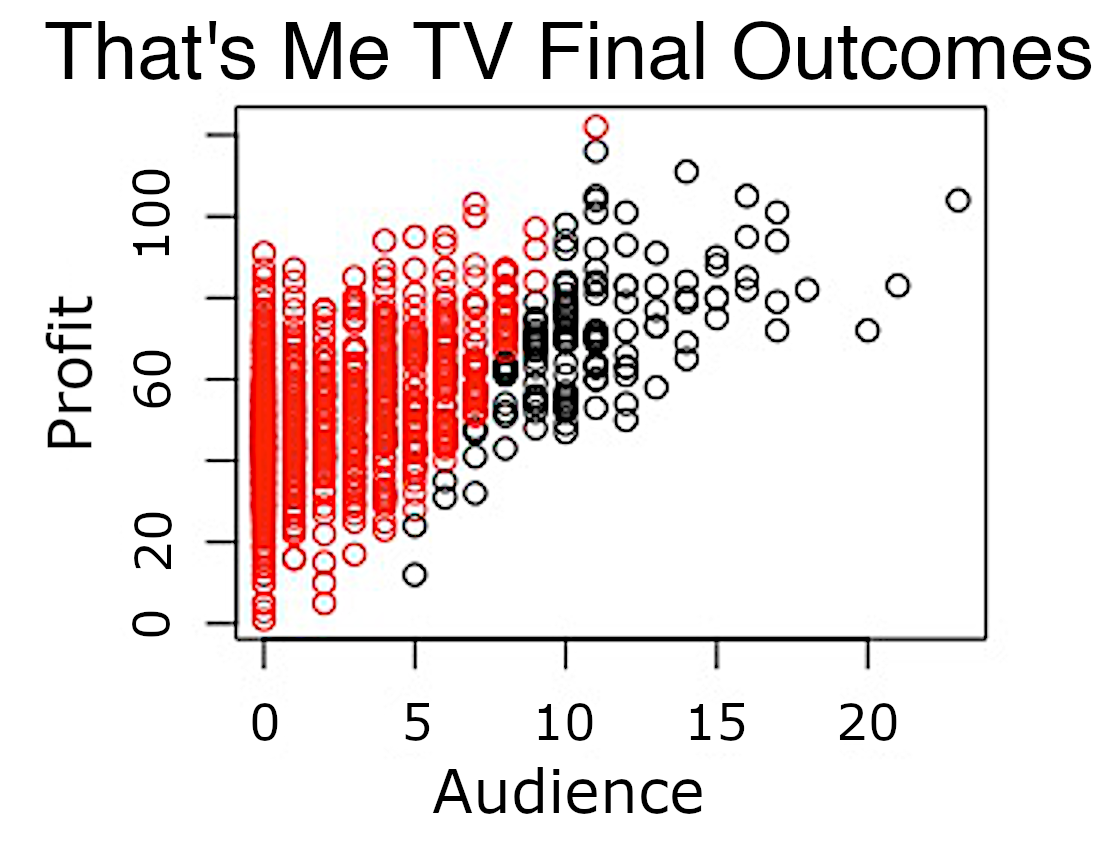}
  \caption{Output expressive range analysis for the game ``That's Me TV'' Red circles indicate ``bad'' endings to the game, while the black circles indicate ``good'' endings.}
  \Description{Output expressive range analysis for the game ``That's Me TV''.}
  \label{fig:expressiveRange}
\end{figure}

Expressive range was originally proposed by Smith and Whitehead \cite{smith2010analyzing} as a means of visualizing the possibility space of a generator. 
It has seen a great deal of interest in PCG research broadly, as both an evaluation \cite{summerville2018expanding} and design tool \cite{cook2016danesh}. 

We argue that expressive range can be researched in TTRPGs as well, which we demonstrate via two examples. 
First, James Malloy employed an approach akin to expressive range during the development of his game \emph{A Space Between}, as is detailed on an episode of the TTRPG design podcast \emph{Stop, Hack, and Roll} \cite{malloy}. %AL: isn't there a citation for this game?
%MJG May 12: Nope, still in development. The podcast is still the best place for it. It's being playtested at cons right now.
\emph{A Space Between} is a game for two players, in which each player draws a number of cards each turn to simulate the lives of a space trucker and someone important to them on Earth. 
As part of his design process, Malloy tagged each of these cards with specific emotions he intended for them to invoke. 
He then was able to lay out each card according to each emotion, essentially visualizing the possibility space of his game.
This allowed him to alter some of the cards and author additional cards, essentially changing his generator to better fit his designer intent \cite{malloy}.
This type of generator iteration, where a designer tags components of the generator in order to better understand its output in terms of these tags, represents an open area for expressive range research. 

Expressive Range was more directly applied in the development of the game \emph{That's Me TV} by Matthew Guzdial \cite{guzdialThats}. 
Given his familiarity with PCG approaches, Guzdial directly applied expressive range for balancing his game. 
In the game, players gather two resources (audience and profit) as they take turns roleplaying as the audience and producers of a children's television show.
Players can take one of a specific set of actions in each role and roll dice to see the effect of the actions. 
Thus, Guzdial was able to simulate out possible end-states for the game, and visualize these as in Figure \ref{fig:expressiveRange}.
Guzdial used this visualization to tweak the mechanics of the game until he found an expressive range that matches his design intention.
This is a much more typical application of expressive range as a design tool, however it represents a novel application domain for the approach.

%AL: confused about this section: so expressive range (a PCG tool) is valuable for TTRPG designers. I thought we were talking about how TTRPG methods/play can be valuable for PCG?
%MJG May 12: Yeah fair! Might need to reframe this section a bit.
%MJG May 16: Updated this section to reflect this.

\subsection{The Fruitful Void}\label{sec:facets_void}%Assigned to: Max + Anne

\begin{figure}[tb]
  \centering
  \includegraphics[width=2.5in]{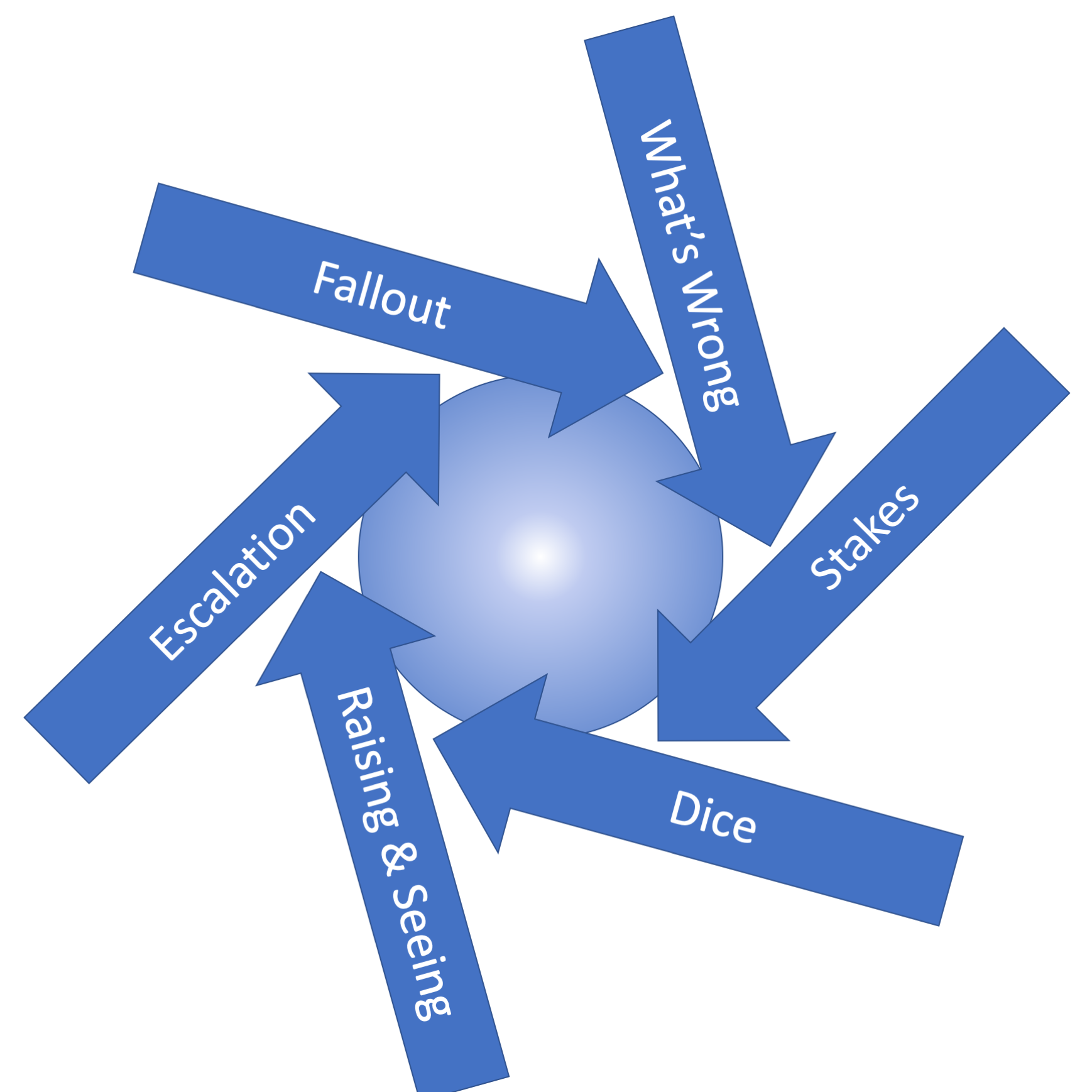}
  \caption{A fruitful void diagram of the tabletop game \emph{Dogs in the Vineyard}, showing how mechanics feed into one another to keep play ``in orbit'' around an unsystematized central theme. Figure based on a figure from \cite{FruitfulVoidThread}.}
  \Description{A recreation of the fruitful void diagram of the tabletop game \emph{Dogs in the Vineyard}.}
  \label{fig:fruitfulVoid}
\end{figure}

In TTRPG design, the term ``fruitful void''~\cite{FruitfulVoidThread} refers to a central theme that is deliberately left unsystematized, but toward which all of the game systems are designed to guide a player's thought and action (see Fig.~\ref{fig:fruitfulVoid}). 
For example, in \emph{My Life with Master} \cite{rpg_lifewithmaster}, ``defiance'' (the game's primary theme) is not systematized directly, but left up to player interpretation, and systematic game mechanics (such as ``fear'', ``weariness'', and ``self-loathing'') are carefully chosen to guide player thoughts \textit{toward} defiance without foreclosing player interpretation. 
One benefit of this approach discussed among TTRPG designers is that players tend to care the most about elements of the story that they had a hand in inventing or shaping, and therefore that a game's central themes tend to come across most strongly when players arrive at an interpretation of these themes for themselves based on play.

%MK: feel free to cut this figure if we need space, i think the visual depiction is helpful for clarity but maybe not necessary

This parallels findings in PCG that indicate a simultaneous need for ambiguity and room for human interpretation in generated artifacts on one hand, and guidance of generation to produce interpretable artifacts on the other. 
Tableau Machine~\cite{TableauMachine}, a system that generates visual art based on the activities of a household's residents, was found to produce intriguingly ambiguous artworks that prompted reflection and extrapolation about the agent's ``state of mind'' among its users. 
An incorrectly installed version of Tableau Machine that generated random artworks without taking real household data into account, however, failed to prompt user reflection -- simultaneously demonstrating the power of ``ambiguity as a resource''~\cite{GaverAmbiguity} and the need for systematic guidance of generative systems that produce ambiguous artifacts. 
Similarly, a study of player storytelling practices around simulation-driven digital games~\cite{EvaluatingViaRetellings} found that player stories are often driven by a process of \emph{extrapolative narrativization}, in which players elaborate on concrete elements of the scenario modeled in the game's systems by adding their own interpretations. This study~\cite{EvaluatingViaRetellings} found that the generative game \emph{Prom Week} \cite{mccoy2013promweek} successfully supports this kind of storytelling by players, but not when the underlying social simulation is altered to produce random outcomes. 
Meanwhile, %PCG researcher and practitioner 
Kate Compton describes one genre of successful Twitter bots as generating ``imagination playgrounds'': strategically ambiguous scenarios with room deliberately left for human interpretation of the ambiguous elements.\footnote{https://twitter.com/GalaxyKate/status/1144758014655655937} 
These examples suggest that designers of PCG systems might benefit from embracing the fruitful void and seeking to produce and present generated artifacts in such a way that encourages a human ``filling in the blanks''. 

\subsection{PCG Pipelines in TTRPGs}\label{sec:facets_pipelines}
%Assigned to: Mike + Devi + Max

PCG pipelines are approaches that tie together several different generators \cite{liapis2019orchestrating}, such that the whole system is able to produce output that no one generator could individually. 
PCG pipelines appear regularly in PCG research, especially in PCGML systems \cite{summerville2018procedural} in which a machine learning-based generator's output is then parsed by a rule-based generator \cite{summerville2016super}%AL: cite example %MJG: done
or its learned space is explored by a search-based generator \cite{volz2018evolving}. %AL: cite example %MJG: done

The same concept of PCG pipelines exists in TTRPGs. 
In some campaigns, one system will be used to generate worlds and histories, while another is actually used to play in that world. 
One example of this is the game \emph{Microscope} \cite{robbins2011microscope}, in which players collaboratively generate a history with different time periods, events, and scenes. 
The end result of this process is a generated history of a world or worlds that players can then use as the setting of a campaign. 
\emph{The Quiet Year}~\cite{rpg_quietyear} can also be used in a similar fashion. 
\textit{Friends at the Table}, a podcast focused on playing TTRPG campaigns, employed \emph{The Quiet Year} to create a city called ``Marielda'', which they used as a setting for a game of \emph{Blades in the Dark} \cite{marielda1}.

Other TTRPGs have world generators built into the game or as a supplement. 
One example of this is the \emph{Engine of Ages}, a collaborative history generator for the game system \emph{13th Age} found in \emph{The Book of Ages} supplement \cite{hanrahan2017ages}. 
In the \emph{Engine of Ages}, players create the history of the world before the campaign of the \emph{13th Age} takes place, describing the history of the previous twelve ages. 
This is done by defining factions and their relationships to one another as they change over time, as well as high and low points in that faction's history. 
These generated elements are based on die rolls that determine during which age and what kind of fact the players are adding to the history of the world. 
This process also determines legends, legacies, and lairs that the players can encounter throughout the course of the campaign, which more closely ties together this world generator game to the main campaign.

Many TTRPGs use generators in their character creation processes through which PCs are constructed. 
The 5th Edition of \emph{D\&D} \cite{rpg_dnd} suggests that players use dice rolls to decide many aspects of a new character, including personality traits and quirks, aspects of the character's personal history, and trinkets they begin a campaign with. 
Character creation generators can be quite layered and complex. 
For example, \emph{Cyberpunk 2020} \cite{rpg_cyberpunk} uses a tree of lookup tables that branch based on earlier dice rolls to build a character background and history, as well as a series of tables which are repeatedly rolled against to construct a year-by-year backstory. 
Lookup tables are a useful structure for player-centric generativity as they are easily operated by players but can be designed to have complex distributions (using techniques such as \emph{Cyberpunk 2020}'s branching).

Another common application of generators to TTRPGs is for determining encounters or obstacles. 
The simplest version of this is some $X{\cdot}Y$ table, pre-authored by a TTRPG designer, where each cell describes an encounter or obstacle for players to face.
These kinds of tables are popular among a branch of TTRPG design referred to as Old School Revival, to the point where there exists a collection of examples playing with this form called \textit{The Strategic Review} \cite{sinclarStrategic}. 
While a simple example of this would be rolling on a random table to see what encounter players face, there are some examples of this with additional complexity. 
For example, the \emph{Mouse Guard RPG}, in which players take on the role of a band of heroic mice, has a rulebook that provides recommendations for creating encounters that involve some combination of threats from weather, wilderness, animals, or other mice \cite{rpg_mouseguard}.
It recommends picking two of those hazards and keeping two on reserve in case of a needed additional twist or challenge. 

Pipelines appear frequently in PCG systems, but are less commonly the topic of PCG research work.
One potential reason for this may be that getting different digital PCG generators to work together can be a challenge \cite{liapis2019orchestrating}, as they likely make use of different programming languages, representations, and code bases.
TTRPGs represent a domain for pipeline research without these challenges, and with many recorded examples to serve as touchstones. 
%AL: This ends a bit abruptly; it would be great if more 'vision-y' stuff could be said here on how PCG pipelines can be enhanced in TTRPGs (e.g. using the more complex knowledge/algorithms used in PCG research)
%MJG: Added above paragraph

\subsection{Digital Mixed-Initiative Agents}\label{sec:facets_agents}%Assigned to: Anne and Mirjam (Devi helping?)
\label{sec:digital-assistance}

As discussed in Section \ref{sec:facets_pipelines}, many TTRPG systems are made up of smaller generators to form the building blocks from which the story is created.
Some of these smaller systems, such as name generators, map generators, loot (treasure) generators, character generators, etc. have been reproduced or augmented with digital tools.
These digital tools can be thought of as equivalent to a more traditional mixed-initiative PCG system. 
In a mixed-initiative PCG system, sometimes called co-creative design, a human designer works together with a digital PCG tool \cite{liapis2016mixedinitiative}.
In this section, we give a number of examples of how these digital aides have been integrated into TTRPGs, though we expect there is a great deal more work to be done in this space.

\emph{Invisible Sun} \cite{rpg_invisiblesun} is a TTRPG that is generally played in a group around a table: it includes a number of physical components such as books, card decks, boards, maps, dice, etc.
To address common issues with getting players in the same physical space, the game developers also created an \emph{Invisible Sun} mobile application\footnote{https://www.montecookgames.com/invisible-sun-app/} to accommodate side scenes for individual or groups of characters that are away from the table and even possibly without the GM.
The application facilitates communication and also provides the players and GM with access to decks that are used for character development play.
Focusing on the generative aspects of the application, the player may draw a random card from one deck to help guide a scene.
However, the application is only providing a digital version of the physical components that are used for storytelling generation.

\emph{Weave} \cite{rpg_weave} is a deck-based storytelling system that, unlike \emph{Invisible Sun}, requires the mobile app to play and therefore the mobile application plays a much large role in story generation\footnote{https://weave.game/}.
The physical components of the game consist of only dice and two decks of tarot-like cards designed with abstracted, symbolic imagery and themes.
In the mobile application, the storyteller can choose a playset for the game, which provides boundaries for the generative space. 
In \emph{Weave}, playsets behave similar to campaign settings although (unlike campaign books in games such as \emph{D\&D}) they do not contain a general story structure. 
The game cards are scanned using the application, which then interprets the card based on the playset chosen by the storyteller, as well as what role the card is playing. 
For instance, during game setup, the storyteller draws a card for the theme of the game. 
If the Gatekeeper card is drawn as the theme, it will provide different themes for each playset, each fitting a generalized interpretation of a gatekeeper. In the playset \emph{Goblins 'R Jerks} where the players take the role of goblins, the Gatekeeper card means the goblins have run out of things to loot, and must find new ways to survive. However, in the playset \emph{Gloomies} which is a 1980s cartoon themed setting, a best friend of one of the players has gone missing and it is up to the party to find out what's going on.

\emph{Spindlewheel} \cite{rpg_spindlewheel}, mentioned in Section \ref{sec:ttrpg_intro}, is a TTRPG system focused on tarot-like cards and interpretation of these cards to create stories. 
A Twitter bot\footnote{\url{https://twitter.com/spindlewheelbot}} for Spindlewheel allows anyone to query for a certain number of these cards, which can then be used to play a Spindlewheel game.
However, all of the official Spindlewheel games can be played without access to this bot, and are typically played in a shared space with physical cards.
The micro-game \emph{How To Build A Place You Love} (a tiny, tweet-sized game) is a Spindlewheel game designed to be played specifically with this bot \cite{okadaHow}. 

We anticipate that applications like these digital tools may be a natural way for PCG researchers to interface with TTRPGs. 
Notably, as in the first and last examples, there's no need to design a new game from scratch to integrate digital PCG. 
Thus, the development of such a companion application for a particular TTRPG game could allow for a broad range of PCG research. 

\section{Discussion}\label{sec:discussion}

In the previous sections we have highlighted a variety of examples showing how concepts and techniques from generative systems design appear in or have been applied to TTRPGs, and vice versa. %AL: I feel like the "and vice versa" is an important addition here, as in some cases we talk about PCG/design assistants within RPGs
%MJG: Agreed! 
The rich history of tabletop roleplaying has many innovative applications of these ideas beyond the examples listed in this paper, and we believe that they have largely been under-explored by digital PCG researchers.

A clear theme throughout our case studies is one of working within constraints. 
For much of the history of TTRPG design, it was not possible to rely on every player having access to a small portable computing device; only recently are we seeing the emergence of digital assistance (see Section \ref{sec:digital-assistance}). 
This means that a lot of classic approaches to generative systems at the table involve what was readily available at the table: objects like dice and coins. 
This restricts the kinds of generative system that can be built, but has also driven innovation among designers to create unpredictable systems that are simple to use. 

The advent of digital aides is an exciting one, as it greatly broadens the opportunities available to researchers and designers. 
However, digital augmentation is not always available to all players, while others prefer to engage with physical objects and non-digital processes. 
Thus, it is important that future research into directly integrating digital PCG considers both digital and physical innovation, or hybrid ways to retain a sense of presence in the physical space while benefiting from the support of digital tools.

The breadth and variety of generative techniques on display here in just a handful of examples that stretch across decades, from niche games to large franchises, show how enduring and popular generative systems are in TTRPGs, and how much work has already been done exploring these ideas and finding new ways to use them for different design goals. There are many exciting possibilities for future work, which should benefit both TTRPG design as well as digital PCG.

\section{Future Work}%Assigned to: Everyone

In this section we present some initial thoughts on potential areas of explicitly PCG research in the domain of TTRPGs, beyond what has already been discussed in this paper.

One fairly straightforward area of future PCG research into TTRPGs would be to replace any of the generative components of a TTRPG with digital PCG components. 
For example, PCG applied to aid GM preparation before or during a session.
We note that one such example already exists: Horswill's FIASCOMATIC \cite{horswill2015fiascomatic} generates playsets for the game \emph{Fiasco}.
However, we expect that significantly more research could be conducted in this area.
We further anticipate that TTRPGs could be a particularly challenging domain for standard PCG approaches, combining story generation and functional ``playability'' constraints. %As noted in Section \ref{sec:digital-assistance}, however, special care must be taken to allow for 

Player modeling is an attempt to automatically understand the players of a digital game \cite{yannakakis2013player}. 
It often comes hand-in-hand with adaptive PCG methods \cite{yannakakis2011experiencedriven} such as dynamic difficulty adjustment \cite{hunicke2005case}.
One could imagine pursuing the equivalent research in TTRPG games, attempting to produce or tweak content on the fly in order to better serve the TTRPG players, for example by ensuring new encounters remain surprising, incorporating the specific histories of the PCs, and so on. 
Importantly, such generators should be able to identify and cater to individual tastes and preferences within the group, e.g. by developing archetypal personas \cite{holmgard2014evolvingpersonas} of different players types such as the ``power gamer''.

Many of the generative approaches used in traditional TTRPGs are grammars or grammar-like in structure (for example, rolling against lookup tables). 
In PCG there is a broader spectrum of techniques used, often partitioned into \textit{search-based} and \textit{constraint-based} generative techniques \cite{togelius2011}. 
Grammars are well-suited for use in physical systems: they usually produce discrete outputs, are built out of similarly discrete components, use simple probabilities and rely on little computation. 
However, attempting to apply different PCG methods to TTRPG design may yield interesting results. 
For example, a TTRPG could be produced in which the entire story was initially purely randomly generated, and then players acted as the search operators over a space of stories, tweaking the initial story to fit their own internal heuristics. 
While state-of-the-art PCG algorithms have been employed for the generation of dungeons \cite{liapis2017multisegment,alvarez2019empowering,ashlock2014automatic}, there is relatively little research on the story-making aspect of TTRPGs.

A common problem faced by digital games researchers is the availability of data, whether that be for statistical purposes such as training models, or simply for manual analysis, criticism and comparison. 
Corpora of useful data such as the Video Game Level Corpus \cite{vglc} can help attract new researchers to an area, makes it easier to compare different research by providing a shared source of baselines or inspiration, and helps pool effort in tracking down hard-to-find data.
For TTRPGs, good data is perhaps even harder to come by. 
Player traces for digital games, for example, can be relatively easily acquired automatically, even when the game is being played remotely, and can be easily parsed into standard formats. 
Transcripts of TTRPGs sessions, by contrast, are very time-consuming and expensive to record, and have many more ethical issues in their acquisition and anonymization. 
The creation of a shared corpus of TTRPG data would therefore be of even greater value than its digital counterpart, and we believe establishing such a corpus will be important in stimulating research in the area.

\section{Conclusions}

This paper that viewing tabletop roleplaying game design as a form of procedural content generator design holds great potential. 
We have introduced TTRPGs for a PCG audience, demonstrated how TTRPGs can be viewed under a PCG lens, and identified relevant areas of PCG research in TTRPGs. 
We believe that PCG research into TTRPGs will allow for a broader understanding of PCG approaches, techniques, and concepts. 

\begin{acks}
Michael Cook is supported by the Royal Academy of Engineering Research Fellowship scheme.
\end{acks}

\bibliographystyle{ACM-Reference-Format}
\bibliography{ttrpg,published_rpgs}

%\end{document}
%\endinput

\appendix

\end{document}